\documentclass[a4paper]{article}

\usepackage{INTERSPEECH2021}
\usepackage{bm}
\usepackage{CJKutf8}
\usepackage{setspace}
\usepackage{subfigure}
\usepackage{threeparttable}
\usepackage{multirow}
\usepackage{cite}
\newcommand{\argmax}{\mathop{\rm arg~max}\limits}

\title{Zero-Shot Joint Modeling of Multiple Spoken-Text-Style Conversion Tasks using Switching Tokens}
\name{Mana Ihori, Naoki Makishima, Tomohiro Tanaka,\\ Akihiko Takashima, Shota Orihashi, Ryo Masumura}
\address{NTT Media Intelligence Laboratories, NTT Corporation, Japan}
\email{mana.ihori.kx@hco.ntt.co.jp}

\begin{document}

\maketitle
\begin{abstract}
  In this paper, we propose a novel spoken-text-style conversion method that can simultaneously execute multiple style conversion modules such as punctuation restoration and disfluency deletion without preparing matched datasets. 
  In practice, transcriptions generated by automatic speech recognition systems are not highly readable because they often include many disfluencies and do not include punctuation marks. 
  To improve their readability, multiple spoken-text-style conversion modules that individually model a single conversion task are cascaded because matched datasets that simultaneously handle multiple conversion tasks are often unavailable. 
  However, the cascading is unstable against the order of tasks because of the chain of conversion errors. 
  Besides, the computation cost of the cascading must be higher than the single conversion. 
  To execute multiple conversion tasks simultaneously without preparing matched datasets, our key idea is to distinguish individual conversion tasks using the {\it on-off switch}. 
  In our proposed zero-shot joint modeling, we switch the individual tasks using multiple switching tokens, enabling us to utilize a zero-shot learning approach to executing simultaneous conversions.
  Our experiments on joint modeling of disfluency deletion and punctuation restoration demonstrate the effectiveness of our method.
\end{abstract}
\noindent\textbf{Index Terms}: spoken text style conversion, zero-shot modeling, switching token

\section{Introduction}

With the rise of various automatic speech recognition (ASR) applications such as smart speakers \cite{googlehome,alexa} and automatic dictation systems \cite{shang2018unsupervised,li2019keep,zhao2019abstractive}, it has become increasingly important to accurately process spoken text that is generated by ASR systems.
However, it is difficult to understand the spoken text because it includes many disfluencies and does not involve punctuation marks.
Moreover, spoken text adversely affects subsequent natural language processing (e.g., machine translation, summarization, etc.) because these technologies are often developed to handle style-converted text that does not include disfluencies and includes punctuation marks.
Therefore, it is important to develop spoken-text-style conversion that converts spoken text into style-converted text.
There have been many studies on spoken-text-style conversion, such as disfluency detection \cite{dong2019adapting}, capitalization \cite{nguyen2019fast}, punctuation restoration \cite{8682418}, and inverse text normalization \cite{pusateri2017mostly}.
Since the advent of deep learning, these studies have achieved high performances by preparing a large amount of data that matches each task.

Although the performance of each conversion task has been improved, multiple spoken-text-style conversions should be handled at the same time to improve the readability of spoken text.
However, preparing a dataset that handles multiple spoken-style-text conversion tasks simultaneously is costly and time-consuming.
Thus, the simplest method of executing multiple tasks simultaneously is to cascade each task.
Unfortunately, the cascading is unstable against the order of tasks because of the chain of conversion errors.
Besides, the computation cost of the cascading must be higher than the single conversion.

\begin{figure}[t]
  \centering
  \centerline{\includegraphics[clip, width=8.0cm]{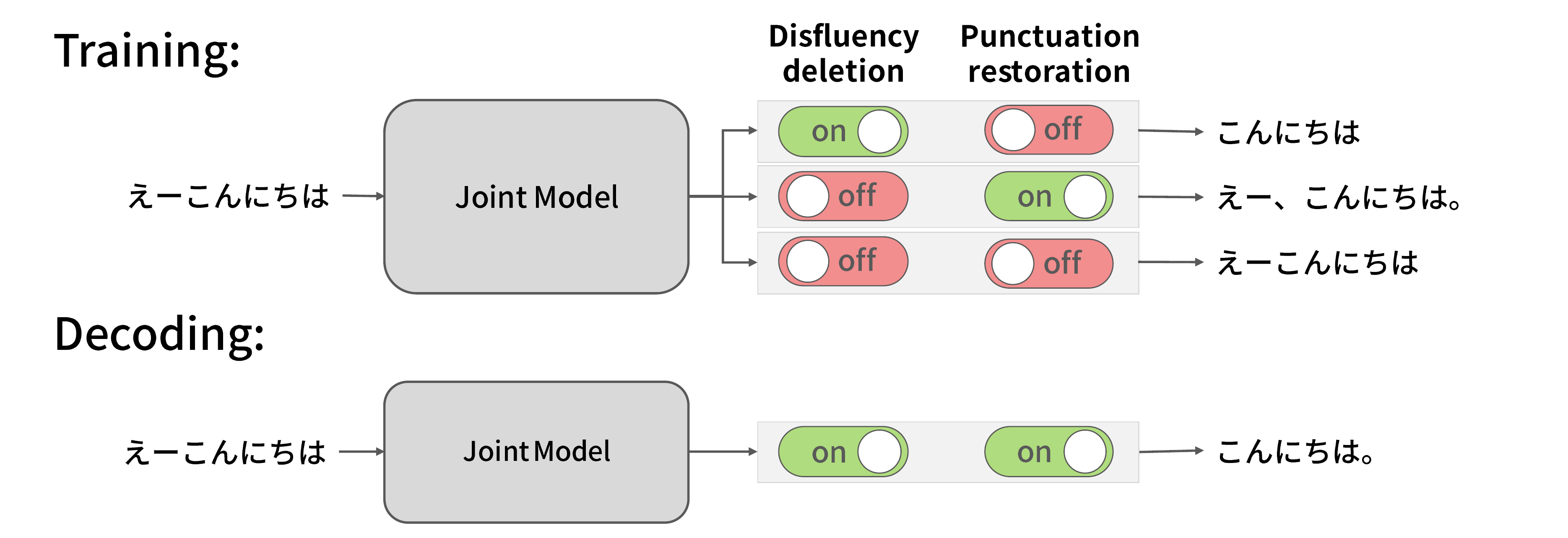}}
  \vspace{-2mm}
  \caption{Zero-shot joint modeling of disfluency deletion and punctuation restoration with the switching.}
  \label{fig:idea}
  \vspace{-6mm}
\end{figure}

To solve these problems, we aim to execute multiple spoken-text-style conversion tasks simultaneously without matched datasets.
Our key idea to execute these tasks simultaneously is to introduce the {\it on-off switch} that can distinguish these individual tasks and a joint conversion task.
In the following, we describe how the model learns disfluency deletion and punctuation restoration at the same time.
Figure \ref{fig:idea} shows an example to execute the disfluency deletion and the punctuation restoration tasks simultaneously.
In Figure \ref{fig:idea}, we prepare the two switches for the disfluency deletion and the punctuation restoration tasks, respectively.
When the model learns the disfluency deletion dataset, it recognizes that disfluency deletion is {\it on} and punctuation restoration is {\it off}.
When the model learns the punctuation restoration, it recognizes that disfluency deletion is {\it off} and punctuation restoration is {\it on}.
Also, when the model learns not to convert (i.e., the input and output text are the same), it recognizes that disfluency deletion and punctuation restoration are both {\it off}.
In this way, all tasks can be recognized in a single model by switching each task {\it on} or {\it off}.
Therefore, the model should execute disfluency deletion and punctuation restoration simultaneously if it switches both tasks {\it on}.
We suppose that even if the model does not learn to execute the joint conversion task at all during training, it can execute multiple conversion tasks simultaneously during inference. 

In this paper, we propose zero-shot joint modeling of multiple spoken-text-style conversion tasks with a Transformer-based encoder-decoder model by using the switching tokens.
The switching tokens switch each conversion task {\it on} or {\it off}.
In our method, the switching tokens are introduced into the decoder for the input contexts.
Our zero-shot joint modeling of multiple conversion tasks starts with two tasks: disfluency deletion and punctuation restoration.
For example, in the disfluency deletion task, two switching tokens $[{\tt disf\_on}]$ and $[{\tt punc\_off}]$ are introduced.
The zero-shot joint modeling is learned using the datasets of disfluency deletion and punctuation restoration tasks simultaneously in a single model with the switching tokens.
Note that our method does not change the architecture of the conversion model, only adds the switching tokens to the model input.
In our experiments using the Corpus of Spontaneous Japanese (CSJ) \cite{maekawa2000spontaneous}, we compare the results of zero-shot modeling using our method with that of cascading.
We demonstrate the effectiveness of the joint modeling of disfluency deletion and punctuation restoration.

\section{Related Work}
\paragraph*{Multiple spoken-text-style conversion tasks:}
Spoken-text-style conversion has multiple tasks.
First, the spoken text has many disfluencies (e.g., fillers and redundant expressions) that are not included in the written text.
The disfluency detection task recognizes non-fluent words in the spoken text \cite{dong2019adapting,Bach2019,tanaka2019,wang2017transition}.
Moreover, disfluency deletion task removes them from the spoken text in an end-to-end manner \cite{ihori2020large}.
Next, capitalization and punctuation are removed from the spoken text because they do not affect the pronunciation.
Therefore, there are tasks that restore capitalization and punctuation for the spoken text \cite{tilk2016bidirectional,nguyen2019fast,8682418,tundik2017bilingual}. 
Moreover, ASR systems output numbers, dates, times, and other numerical entities in a literal manner (e.g., 20 $\rightarrow$ twenty).
Thus, inverse text normalization was proposed to format entities like numbers, dates, times, and addresses \cite{ju2008language,pusateri2017mostly,sunkara2021neural}.
In addition, a neural generation model of ASR spelling correction \cite{guo2019spelling} and grammar correction \cite{hrinchuk2020correction} was proposed to deal with the many spelling and grammar errors in the spoken text.
In this way, the spoken text involves many tasks to be converted, but these tasks have been studied individually.

\paragraph*{Spoken-to-written style conversion task:}
A few studies have been focusing on handling multiple spoken-text-style conversion tasks at the same time.
ASR post-processing for readability (APR) was proposed in \cite{liao2020improving}.
APR is the task of converting spoken-style text into written-style text, and this method converts capitalization, disfluency, and grammar errors, as well as properly formatted dates, times, and other numerical entities simultaneously. 
In this previous study, the dataset to convert these entities simultaneously is prepared using text-to-speech and ASR systems.
Moreover, spoken-to-written style conversion was proposed in \cite{ihori2020parallel,ihori-etal-2020-memory,ihori2021mapgn}.
Spoken-to-written style conversion is the task of converting style unification, postpositional particle expressions restoration, notation correction, punctuation restoration, disfluency deletion, simplification, error correction simultaneously.
In this previous study, the dataset is prepared using crowdsourcing.
These previous studies need a dataset that can handle multiple tasks simultaneously, but such datasets are not published officially, and it is costly and time-consuming to prepare one.

\section{Spoken-Text-Style Conversion with Transformer}
This section defines spoken-text-style conversion.
In this paper, we model spoken-text-style conversion as a sequence-to-sequence problem in which the source is spoken text and the target is style-converted text.
Moreover, we utilize Transformer-based encoder-decoder network architecture \cite{vaswani2017attention}.
We define the spoken text as $\bm{X} = \{x_1, \cdots, x_m\}$ and the style-converted text as $\bm{Y} = \{y_1, \cdots, y_n\}$, where $x_m$ is a token in spoken text and $y_n$ is a token in style-converted text.
The Transformer predicts the generation probabilities of a style-converted text $\bm{Y}$ given a spoken text $\bm{X}$.
The generation probability of $\bm{Y}$ is defined as
\begin{equation}
  P(\bm{Y}|\bm{X}; \bm{\Theta}) = \prod_{n=1}^N P(y_n|y_{1:n-1}, \bm{X}; \bm{\Theta}) ,
\end{equation}
where $\bm{\Theta}$ represents model parameter sets.
% $\theta_{\rm{enc}}$ and $\theta_{\rm{dec}}$ are trainable parameter sets with an encoder and a decoder, respectively.
$P(y_n|y_{1:n-1},$ $\bm{X};$ $\bm{\Theta})$ can be computed with the Transformer encoder and the Transformer decoder.

\paragraph*{Training:}
We treat the disfluency deletion and punctuation restoration tasks as the spoken-text-style conversion tasks.
The disfluency deletion model is trained using a disfluency deletion dataset ${\cal D}_{\rm disf}$.
The training loss function ${\cal L}_{\rm disf}$ is defined as
\begin{equation}
  {\cal L}_{\rm disf} = - \sum_{(\bm{X},\bm{Y}) \in {\cal D}_{\rm disf}} \log P(\bm{Y}|\bm{X}; \bm{\Theta}_{\rm disf}). 
\end{equation}
Moreover, the punctuation restoration model is trained using a punctuation restoration dataset ${\cal D}_{\rm punc}$.
The training loss function ${\cal L}_{\rm punc}$ is defined as
\begin{equation}
  {\cal L}_{\rm punc} = - \sum_{(\bm{X},\bm{Y}) \in {\cal D}_{\rm punc}} \log P(\bm{Y}|\bm{X}; \bm{\Theta}_{\rm punc}). 
\end{equation}

\paragraph*{Cascading decoding:}
In this paper, we aim to execute a disfluency deletion task and a punctuation restoration task at the same time.
When individual conversion tasks are independently modeled from individual datasets, we can execute the joint conversion task via cascading decoding.
The decoding problem in the cascading of disfluency deletion and punctuation restoration is defined as
\begin{equation}
  \tilde{\bm{Y}} = \argmax_{\bm{Y}} P(\bm{Y}|\bm{X}, \bm{\Theta}_{\rm disf}),
\end{equation}
\begin{equation}
  \hat{\bm{Y}} = \argmax_{\bm{Y}} P(\bm{Y}|\tilde{\bm{Y}} , \bm{\Theta}_{\rm punc}).
\end{equation}
Here, the order of cascading can also be reversed.
Note that the conversion performance is varied by the order of cascading.

\section{Proposed Method}
\subsection{Strategy}
\begin{figure}[t]
  \centering
  \centerline{\includegraphics[clip, width=8.0cm]{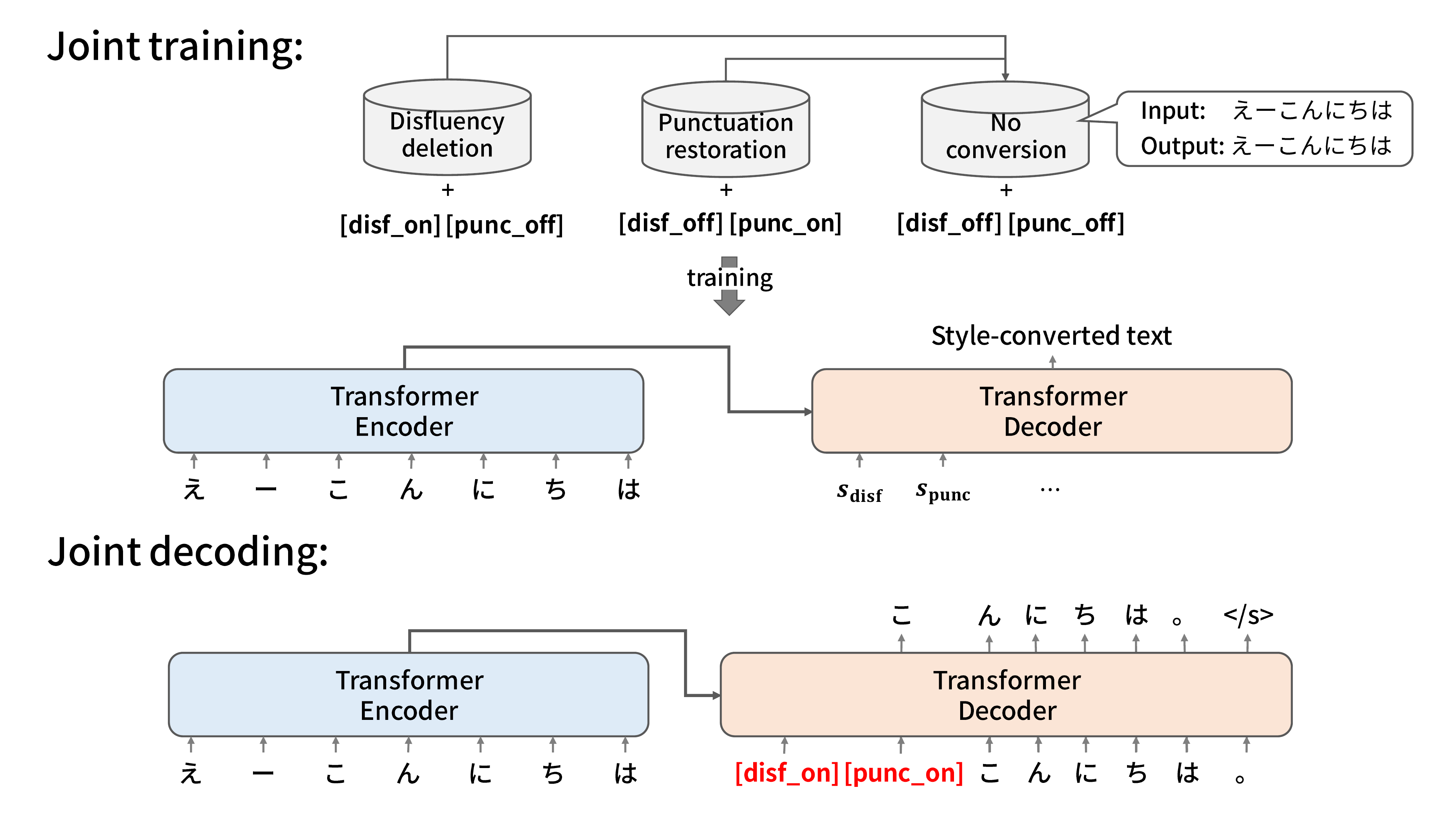}}
  \vspace{-2mm}
  \caption{Proposed method of jointly modeling multiple spoken-text-style conversions.}
  \label{fig:proposed}
  \vspace{-7mm}
\end{figure}

We utilize switching tokens to model multiple spoken-text-style conversion tasks simultaneously without needing to prepare a matched dataset.
Figure \ref{fig:proposed} shows multiple spoken-text-style conversion with Transformer-based encoder-decoder model using the switching tokens.
The switching tokens distinguish tasks and represent the {\it on} state (the target for conversion) or {\it off} state (not the target for conversion) in each task.
For example, if the model is trained with a disfluency deletion dataset, two switching tokens $[{\tt disf\_on}]$ and $[{\tt punc\_off}]$ are given.
Also, if the model is trained with a punctuation restoration dataset, two switching tokens $[{\tt disf\_off}]$ and $[{\tt punc\_on}]$ are given.
Here, ``disf'' represents disfluency deletion and ``punc'' represents punctuation restoration.

In this way, even if the model does not train the state of  $[{\tt disf\_on}]$ $[{\tt punc\_on}]$, it should be able to convert two conversion tasks simultaneously during inference. 
However, the model trained with these two switching patterns ($[{\tt disf\_on}]$ $[{\tt punc\_off}]$ and $[{\tt disf\_off}]$ $[{\tt punc\_on}]$) may be convinced that one task must be {\it off} if the other task is {\it on}.
Thus, we suppose that the model trained with only these two switching patterns cannot handle the joint conversion task at all.
To make the model learn other states of $ [{\tt disf\_on}]$ $[{\tt punc\_off}] $ and $ [{\tt disf\_off}]$ $[{\tt punc\_on}]$, we make a dataset of the $[{\tt disf\_off}]$ $[{\tt punc\_off}]$ state.
This ``no conversion'' dataset can be constructed from the disfluency deletion and the punctuation restoration datasets by supposing the input text is not changed by the conversion.
% The no conversion dataset is made by using input text in punctuation restoration dataset and disfluency deletion dataset.
We expect that the model understands that one task is not necessarily {\it off} if the other task is {\it on} by training with these three datasets.
Even if we combine switching tokens that are not trained to the model, we suppose the model can generate the text because it understands the relation of each task and the switching tokens.

\subsection{Joint Modeling}
The switching tokens are introduced into the decoder for the input contexts, and our model handles disfluency deletion, punctuation restoration, no conversion, and joint conversion tasks as an all-in-one modeling.
In other words, our method can learn multiple tasks simultaneously simply by giving the model the switching tokens without changing the model architecture.
In spoken-text-style conversion using the switching tokens as context, the generation probability of $\bm{Y}$ is defined as
\begin{multline}
  P(\bm{Y}|\bm{X}, s_{\rm disf}, s_{\rm punc}; \bm{\Theta}) = \\
  \prod_{n=1}^N P(y_n|y_{1:n-1}, \bm{X}, s_{\rm disf}, s_{\rm punc}; \bm{\Theta}),
\end{multline}
\begin{equation}
  s_{\rm dist} \in \{[{\tt disf\_on}], [{\tt disf\_off}]\},
\end{equation}
\begin{equation}
  s_{\rm punc} \in \{[{\tt punc\_on}], [{\tt punc\_off}]\}.
\end{equation}
Here, if we set $[{\tt disf\_on}]$ $[{\tt punc\_off}]$, $[{\tt disf\_off}]$ $[{\tt punc\_on}]$, and $[{\tt disf\_off}]$ $[{\tt punc\_off}]$, the model execute the disfluency deletion task, the punctuation restoration task, and the no conversion task, respectively.
Moreover, if we set $[{\tt disf\_on}]$ $[{\tt punc\_on}]$, the model execute the disfluency deletion task and the punctuation restoration task jointly.

\paragraph*{Joint training:}
In our method, the disfluency deletion dataset ${\cal D}_{\rm disf}$, the punctuation restoration dataset ${\cal D}_{\rm punc}$, and the no conversion dataset ${\cal D}_{\rm same}$ are trained jointly in a single model.
The no conversion dataset is made by using the input text of the disfluency deletion dataset and the punctuation dataset.
Thus, the model is trained with the sum of twice the amount of disfluency deletion dataset and that of the punctuation restoration dataset.
Note that the model is not trained with a matched dataset for the disfluency deletion and punctuation restoration.
The training loss function ${\cal L}$ is defined as
\begin{equation}
  {\cal L} = {\cal L}_{\rm disf} + {\cal L}_{\rm punc} + {\cal L}_{\rm same},
\end{equation}
\begin{multline}
  {\cal L}_{\rm disf} = - \sum_{(\bm{X},\bm{Y}) \in {\cal D}_{\rm disf}} \log P(\bm{Y}|\bm{X}, \\
  s_{\rm disf} = [{\tt disf\_on}], s_{\rm punc} = [{\tt punc\_off}]; \bm{\Theta}),
\end{multline}
\begin{multline}
  {\cal L}_{\rm punc} = - \sum_{(\bm{X},\bm{Y}) \in {\cal D}_{\rm punc}} \log P(\bm{Y}|\bm{X}, \\
  s_{\rm disf} = [{\tt disf\_off}], s_{\rm punc} = [{\tt punc\_on}]; \bm{\Theta}),
\end{multline}
\begin{multline}
  {\cal L}_{\rm same} = - \sum_{(\bm{X},\bm{Y}) \in {\cal D}_{\rm same}} \log P(\bm{Y}|\bm{X}, \\
  s_{\rm disf} = [{\tt disf\_off}], s_{\rm punc} = [{\tt punc\_off}]; \bm{\Theta}).
\end{multline}

\paragraph*{Joint decoding:}
When the disfluency deletion and punctuation restoration tasks are performed simultaneously, we give the $[{\tt disf\_on}] [{\tt punc\_on}]$ switching tokens to the model.
The decoding problem of zero-shot modeling using the switching tokens is defined as
\begin{multline}
  \hat{\bm{Y}} = \argmax_{\bm{Y}} P(\bm{X}|\bm{Y}, \\
  s_{\rm disf} = [{\tt disf\_on}], s_{\rm punc} = [{\tt punc\_on}]; \bm{\Theta}).
\end{multline}
Here, if we feed the switching tokens $[{\tt disf\_on}]$ and $[{\tt punc\_off}]$ into the model, it performs the disfluency deletion task only.
Moreover, if we feed the switching tokens $[{\tt disf\_off}]$ and $[{\tt punc\_on}]$ into the model, it performs the punctuation restoration task only.

\section{Experiments}
\vspace{-1mm}
\subsection{Datasets}
We evaluated effectiveness of our method using the CSJ dataset \cite{maekawa2000spontaneous}.
We divided the CSJ into a training set, a validation set, and a test set.
We used disfluency deletion and punctuation restoration for the spoken-text-style conversion task.
Thus, we divided a training set and a validation set for each task.
First, in the punctuation restoration task, we made pair data of the spoken text and the text with punctuation marks restored using crowdsourcing.
We prepared a 50,000-sentence training set and a 5,000-sentence validation set for the punctuation restoration task.
Next, in the disfluency deletion task, we made pair data of the spoken text and the text with fillers removed based on the part of speech in the CSJ.
Here, we prepared a 50,000-sentence training set and a 5,000-sentence validation set to prevent imbalanced datasets from affecting the conversion performance.
% Finally, we prepared a test sets: pair data of spoken text and the text with fillers removed, the text with punctuation marks restored, and the text with fillers removed and punctuation marks restored.
% The test set had 3,949 sentences.
Finally, we prepared a test set for each task and it consists of 3,949 sentences.
In addition, we investigated how performance differed depending on the amount of data in the training set by dividing the training set into 10,000, 30,000, and 50,000 sentences, respectively.

\begin{table*}[t]
  \scriptsize
  \centering
  \caption{\label{table:each_task} Results of disfluency deletion and punctuation restoration, respectively.}
  \vspace{-2mm}
  \begin{threeparttable}
  \begin{tabular}{|rcc|rrr|rrr|rrr|} 
      \hline
       & \multicolumn{2}{c|}{Task} & \multicolumn{3}{|c|}{10k sentences in each dataset} & \multicolumn{3}{|c|}{30k sentences in each dataset} & \multicolumn{3}{|c|}{50k sentences in each dataset}\\
       & disf & punc & BLEU & METEOR & GLEU & BLEU & METEOR & GLEU & BLEU & METEOR & GLEU \\
       \hline \hline
       1). & \checkmark & - & 0.879 & 0.981 & 0.831 & 0.901 & 0.988 & 0.858 & 0.909 & 0.991 & 0.870 \\
       3). & on & off & 0.892 & 0.985 & 0.849 & 0.902 & 0.987 & 0.864 & 0.905 & 0.990 & 0.864 \\
       \hline \hline
       2). & - & \checkmark & 0.767 & 0.945 & 0.644 & 0.809 & 0.962 & 0.693 & 0.816 & 0.964 & 0.700 \\
       3). & off & on & 0.838 & 0.981 & 0.725 & 0.839 & 0.982 & 0.727 & 0.844 & 0.983 & 0.735 \\
      \hline 
  \end{tabular}
  \begin{tablenotes} \footnotesize
    \item 1). disfluency deletion model \hspace{2mm} 2). punctuation restoration model \hspace{2mm} 3). switching token-based joint model
  \end{tablenotes} 
\end{threeparttable}
\vspace{-2mm}
\end{table*}

\begin{table*}[t]
  \scriptsize
  \centering
  \caption{\label{table:multiple_tasks} Results of multiple spoken-text-style conversion tasks.}
  \vspace{-2mm}
  \begin{threeparttable}
  \begin{tabular}{|rcc|c|c|rrr|rrr|rrr|} 
    \hline
     &\multicolumn{2}{c|}{Task} & \multirow{2}{*}{Speed} & \multirow{2}{*}{Memory} & \multicolumn{3}{|c|}{10k sentences in each dataset} & \multicolumn{3}{|c|}{30k sentences in each dataset} & \multicolumn{3}{|c|}{50k sentences in each dataset}\\
     & disf & punc & & & BLEU & METEOR & GLEU & BLEU & METEOR & GLEU & BLEU & METEOR & GLEU \\
     \hline \hline
     \multirow{2}{*}{1+2).} & \multicolumn{2}{c|}{cascading ($\rightarrow$)} & $\times$ 2.0 & $\times$ 2.0 & 0.765 & 0.944 & 0.668 & 0.824 & 0.969 & 0.730 & 0.839 & 0.974 & 0.757 \\
      & \multicolumn{2}{c|}{cascading ($\leftarrow$)} & $\times$ 2.0 & $\times$ 2.0 & 0.683 & 0.889 & 0.545 & 0.750 & 0.923 & 0.622 & 0.764 & 0.929 & 0.640 \\

     \hline \hline
     \multirow{3}{*}{3).} & \multicolumn{2}{c|}{cascading ($\rightarrow$)} & $\times$ 2.0 & $\times$ 1.0 & 0.805 & 0.972 & 0.712 & 0.815 & 0.974 & 0.730 & 0.820 & 0.977 & 0.738 \\
      & \multicolumn{2}{c|}{cascading ($\leftarrow$)} & $\times$ 2.0 & $\times$ 1.0 & 0.722 & 0.898 & 0.596 & 0.738 & 0.904 & 0.609 & 0.733 & 0.905 & 0.608 \\
     \cline{2-14}
      & on & on & $\times$ 1.0 & $\times$ 1.0 & 0.799 & 0.972 & 0.704 & 0.814 & 0.974 & 0.726 & 0.818 & 0.973 & 0.725 \\
      \hline 
  \end{tabular}
  \begin{tablenotes} \footnotesize
    \item 1). disfluency deletion model \hspace{2mm} 2). punctuation restoration model \hspace{2mm} 3). switching token-based joint model
  \end{tablenotes} 
  \end{threeparttable}
  \vspace{-6mm}
\end{table*}

\vspace{-1mm}
\subsection{Setups}
For evaluation purposes, we constructed three Transformer based encoder-decoder models: 1) a disfluency deletion model, 2) a punctuation restoration model, and 3) a switching token-based joint model.
The disfluency deletion and punctuation restoration models were trained using only each dataset without the switching tokens.
The switching token-based joint model was our proposed model trained with both datasets.
Note that the Transformer architecture was the same in these models.

We employed the following configurations.
In the encoder, a 4-layer Transformer encoder block with 512 units was introduced.
In the decoder, a 2-layer Transformer decoder block with 512 units was introduced.
The output unit size (corresponding to the number of tokens in the training and validation sets) was set to 3,316.
% The output unit size was set to 3,316.
To train the Transformer, we used the RAdam optimizer \cite{liu2019variance} and label smoothing \cite{pmlr-v119-lukasik20a} with a smoothing parameter of 0.1.
We set the mini-batch size to 64 sentences and the dropout rate in each Transformer block to 0.1.
% For the mini-batch training, we truncated each sentence to 200 tokens.
All trainable parameters were randomly initialized, and we used characters as tokens.
For the decoding, we used a beam search algorithm in which the beam size was set to 4.
For the evaluation, we calculated the automatic evaluation scores using three metrics: BLEU \cite{papineni2002bleu}, METEOR \cite{banerjee-lavie-2005-meteor}, and GLEU \cite{napoles-etal-2015-ground}.
BLEU and METEOR are metrics used in machine translation tasks, and they compute the matching of the reference and hypothesis text.
GLEU is a metric used in grammatical error correction tasks \cite{napoles-etal-2019-enabling}, and it is computed using the input, reference, and hypothesis text.
If the model generates n-gram that is not in the input text but is in the reference text, the evaluation score becomes high.
In this paper, we use 4-gram for BLEU and GLEU.

\vspace{-2mm}
\subsection{Results}
Table \ref{table:each_task} shows the respective evaluation results of the disfluency deletion task and the punctuation restoration task.
Moreover, Table \ref{table:multiple_tasks} shows the evaluation results of cascading decoding with the disfluency deletion and punctuation restoration models, cascading decoding with switching token-based joint model, and the joint decoding with the joint model.
Cascading ($\rightarrow$) in Table \ref{table:multiple_tasks} represents the order of cascading, and cascading ($\rightarrow$) was cascaded in the order of the disfluency deletion task to the punctuation restoration task.
Cascading ($\leftarrow$) was processed in the reverse order.
In Table \ref{table:multiple_tasks}, the joint modeling where each switch is {\it on} is our proposed zero-shot joint decoding.

First, we focus on the results in Table \ref{table:each_task}.
In the disfluency deletion task, when the amount of training data was 50,000 sentences, the disfluency deletion model slightly outperformed our method.
Otherwise, our method outperformed the disfluency deletion model.
Moreover, in the punctuation restoration task, our method substantially outperformed the the punctuation restoration model.
This indicates that joint modeling using datasets of all tasks improves the performance of each task.
Next, we focus on the results in Table \ref{table:multiple_tasks}.
The cascading results show that the performance was substantially different based on the cascading order in all models.
The result of cascading ($\rightarrow$) substantially outperformed the result of cascading ($\leftarrow$).
The reason is that the model replaced punctuation marks that were generated in the punctuation restoration task with other words because it does not learn to generate punctuation marks in the disfluency deletion task.
Here, the result of cascading ($\leftarrow$) using our method substantially underperformed the result using baseline because our method has high performance of punctuation restoration task and many punctuation marks were replaced with other words by cascading.
Therefore, in cascading, it is difficult to convert spoken text robust because the order of processing has a large effect on performance.

In addition, we focus on the results of the zero-shot joint decoding in Table \ref{table:multiple_tasks}.
When the training data was 10,000 sentences, the zero-shot joint decoding outperformed the cascading decoding with the disfluency deletion and punctuation restoration models.
Moreover, when the training data was 10,000 and 30,000 sentences, the zero-shot joint decoding performance was comparable with cascading decoding with switching token-based joint model.
On the other hand, when there were 50,000 sentences in each dataset, the cascading ($\rightarrow$) with the disfluency deletion and punctuation restoration models slightly outperformed the zero-shot joint decoding because influence against chain of conversion errors was mitigated.
This indicates that if the amount of data in each dataset is small, the performance of the zero-shot joint decoding is effective to prevent the chain of conversion errors.
% Also, the advantage of our method is that the order of processing does not matter as it does for cascading.
% Although the zero-shot conversion using our method slightly underperforms the baseline, it can convert multiple spoken-text-style conversion with half the parameters and computation cost. 
These results confirm that our proposed zero-shot joint decoding with switching token joint modeling are stable approach and can achieve comparable performance to cascading decoding with the best choice of the processing order although matched datasets for the joint conversion problem are unavailable.
Moreover, the proposed joint decoding is twice as fast as the cascading decoding while keeping the superior performance.

% \vspace{-2mm}
\section{Conclusions}
% \vspace{-1mm}
In this paper, we proposed a spoken text conversion method that can simultaneously execute multiple style conversion modules without needing to prepare a matched dataset.
Our key idea is to distinguish individual conversion tasks using the {\it on-off switch} with multiple switching tokens.
The switching tokens enable us to utilize a zero-shot learning approach to executing simultaneous conversions.
In the evaluation experiment, we treated disfluency deletion and punctuation restoration tasks as multiple spoken-text-style conversion tasks.
The results demonstrated that our method can execute multiple conversion tasks simultaneously that is twice as fast as the cascading while keeping superior performance even though matched training dataset was not used at all.
% In future work, we will increase the number of tasks to train with a single model and apply the pre-training for the spoken-text-style conversion model.

\clearpage
% {
%   \bibliographystyle{IEEEtran}
%   \bibliography{mybib}
% }
{

}
\end{document}